\documentclass[conference]{IEEEtran}
\IEEEoverridecommandlockouts
\usepackage{cite}
\usepackage{amsmath,amssymb,amsfonts}
\usepackage{algorithmic}
\usepackage{graphicx}
\usepackage{textcomp}
\usepackage{xcolor}
\usepackage{graphicx}
\usepackage{subcaption}
\def\BibTeX{{\rm B\kern-.05em{\sc i\kern-.025em b}\kern-.08em
    T\kern-.1667em\lower.7ex\hbox{E}\kern-.125emX}}
\usepackage{fancyhdr}
\definecolor{red}{rgb}{1.0,0.0,0.0}
\definecolor{orange}{rgb}{1.0,0.65,0.0}
\definecolor{blue}{rgb}{0.0, 0.0, 1.0}

\newcommand{\fig}[1] {Fig.~\ref{#1}}

\begin{document}

\title{Mind in the Machine: 
Perceived Minds Induce Decision Change\\
}

\author{\IEEEauthorblockN{Deniz Lefkeli, }
\and
\IEEEauthorblockN{Baris Akgun, }
\and
\IEEEauthorblockN{Sahibzada Omar, }
\and
\IEEEauthorblockN{Aansa Malik, }
\and
\IEEEauthorblockN{Zeynep Gurhan Canli, }
\and
\IEEEauthorblockN{Terry Eskenazi}
}

\maketitle
\thispagestyle{fancy}

\begin{abstract}
Recent research on human robot interaction explored whether people's tendency to conform to others extends to artificial agents \cite{hertz2016influence}. However, little is known about to what extent perception of a robot as having a mind affects people’s decisions. Grounded on the theory of mind perception, the current study proposes that artificial agents can induce decision change to the extent in which individuals perceive them as having minds. By varying the degree to which robots expressed ability to act (agency) or feel (experience), we specifically investigated the underlying mechanisms of mind attribution to robots and social influence. Our results show an interactive effect of perceived experience and perceived agency on social influence induced by artificial agents. The findings provide preliminary insights regarding autonomous robots’ influence on individuals’ decisions and form a basis for understanding the underlying dynamics of decision making with robots.
\end{abstract}


\section{Introduction} \label{sec:intro}
In the last decade, studies on the nature and consequences of human-robot interaction gained traction providing theoretical and practical insights for making such interactions as smooth and successful as possible \cite{broadbent2017interactions}. Mind perception, which is perceiving the mere presence of mind in another entity such as humans, animals, or technological devices, provide a perspective that can be useful in understanding the dynamics of human robot interaction \cite{epley2010mind}. Attributing mental states such as planning, reasoning, emotion, desire and consciousness enables prediction and interpretation of others’ desires \cite{epley2004perspective, keysar2000taking}, beliefs \cite{hare2002domestication}, and behaviors \cite{epley2008we, kozak2006think}. This process starts with determining whether an entity has a mind or not, but different entities may show variance in the dimensions of mind \cite{gray2007dimensions}. Depending on the perceived level of these dimensions, an entity can be considered as an agent that has a mind capable of experiencing complex emotions or a limited capacity of processing \cite{kozak2006think}.

\begin{figure}
    \centering
    \includegraphics[width=0.99\columnwidth]{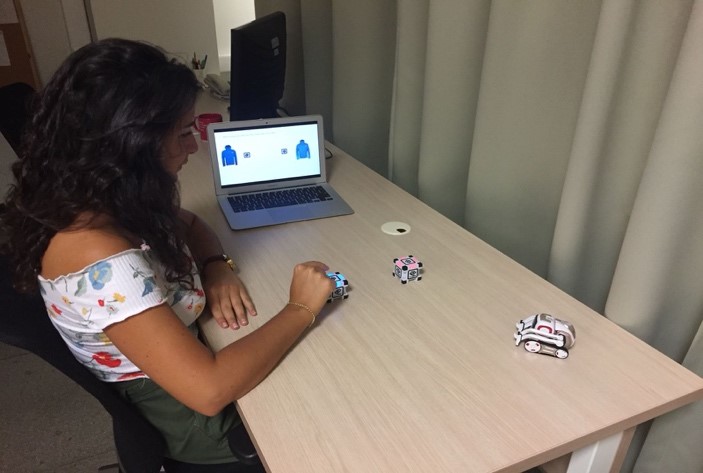}
    \caption{Participants make subjective and objective choices and receive recommendations from Cozmo. After Cozmo's suggestions, participants have a chance to change their decision.}
    \label{fig:part_cozmo}
\end{figure}

Research on human robot interaction has shown that robots can induce behavior change in ambiguous tasks \cite{hertz2016influence}. However, how people’s objective and subjective decisions change upon interacting with a robot that is perceived as having a mind has not been studied before. Exploring social influence in the context of human robot interaction, we specifically investigated the underlying mechanisms of the link between mind perception and decision change, and whether the variability in dimensions of mind influences different types of decisions in a real-life setting. Through four experimental conditions, we orthogonally manipulated perceived agency and perceived experience of a robot which provided suggestions to a participant on two types of questions. Subjective questions asked the participants for their preference between two items (\fig{fig:tasks}). Objective questions were mathematical in nature and had a correct answer. We were interested in participants’ tendency to change their initial decisions upon the suggestion given by the robot. Although prior research has explored conformity in human-robot interaction, this study differentiates itself in the social robotics field as it grounds on a consumer context which provides an opportunity to observe human-robot interaction in a setting that is expected to become a part of daily life. By providing insights about decision making processes in the context of human robot interaction setting, findings of this study will contribute to the prior literature on robotics and social influence.

\section{RELATED WORK} \label{sec:rel-work}

\subsection{Conforming to Robots} 
Conformity, as one of the powerful aspects of social influence, is broadly defined as a change in behavior in order to match with the group norm \cite{cialdini2004social,bailenson2008use}. People’s tendency to conform to the rest of the group’s decision can be driven by different motivations \cite{deutsch1955study}. Epistemic (informational) conformity refers to complying to others because of one’s uncertainty about their own decisions \cite{deutsch1955study}. Task ambiguity and perceived knowledge level of the group may thus influence individuals’ tendency to conform. Normative conformity refers to conforming to the expectations of other people, groups, or even one’s own self due to the experienced peer pressure \cite{deutsch1955study}. It is a change in behavior to be accepted by the group and avoid rejection. Research has shown that people tend to switch their answers to match with those of the group even when they think that the group’s answer is wrong \cite{asch1956studies}. 

Recent studies have explored whether conformity extends to artificial agents. Studies applying well known conformity paradigms with robots \cite{shiomi2013recommendation, hertz2016influence, brandstetter2014peer} found that people are more likely to conform to the majority decision of individuals than to the majority decision of virtual agents \cite{midden2015conforming}.  Research has also shown that the tendency to conform to an agent, which is presented as a computer, a human, or a robot, does not always differ across agent types; rather, the difficulty level of the task can increase the level of conformity in objective tasks where questions have correct answers \cite{hertz2016influence}. Another study found that virtual humans can make people conform; even though they failed to see an effect of agency (individual’s belief that virtual humans are represented by real people) or behavioral realism (realistic representation of the human behavior) on conformity levels \cite{kyrlitsias2018asch}. Investigating people’s tendency to conform with artificial agents in various tasks, these studies have shown that task difficulty and perceived competence in the task, rather than the agent type, increases people’s conformity levels. Lack of a social connection between humans and robots, varying levels of anthropomorphization, and the lack of an authoritative impression of a robot have been suggested as the possible reasons for the inability to observe humans’ conformity to robots \cite{shiomi2016synchronized,brandstetter2014peer,salomons2018humans}. 
In line with the recent findings in human-robot interaction research that focuses on conformity, the proposed study explores whether robots are capable of changing people’s behavior in both subjective and objective tasks.

\subsection{Mind Perception}
Mind perception can be defined as the comprehension of the mere presence of mind in other entities, including humans, animals, or artificially intelligent agents \cite{epley2010mind}. It facilitates predicting and interpreting others’ beliefs, desires, and behaviors \cite{epley2004perspective,epley2006perspective,kozak2006think}. Research has suggested that people perceive minds on two dimensions: agency and experience \cite{gray2007dimensions}. Agency refers to the capacity to plan and act that enable decision-making and organization of behaviors, which covers the capacities of self-control, morality, memory, emotion recognition, planning, communication and thought; while experience refers to the capacity to sense and feel, which can be attributed from hunger, fear, pain, pleasure, rage, desire, personality, consciousness, pride, embarrassment and joy \cite{gray2007dimensions}. 

However, there is a variance in the degree of mind perception \cite{gray2007dimensions}. Agents that are high on the mind scale can experience complex emotions and contemplate on complex ideas, while agents that are low on the mind scale have limited capabilities of cognition and experience \cite{morewedge2007timescale}. Further, living and non-living entities are expected to be perceived differently in terms of their dimensions of mind. While adult humans are attributed both experience and agency, children and animals are perceived as having higher capacities for experience than agency \cite{schein2015unifying}. 

Recent theoretical accounts emphasize the uniqueness of the experience dimension as robots and other artificial agents are considered as being less capable of experiencing emotions and having sensations \cite{gray2007dimensions,huebner2010commonsense,knobe2008intuitions,schein2015unifying}. Consequently, many people consider robots as having some component of mind, especially agency, which suggests that exploring the variance in mind perception is important \cite{broadbent2017interactions}.

Research has suggested that people treat machines as if they were capable of experiencing \cite{epley2007seeing,waytz2010social}, and they actually prefer a virtual agent that can express its emotions to a neutral one \cite{creed2015impact}. However, a computer without a human-like appearance that can experience can create a sense of uneasiness in humans \cite{gray2012feeling}. Similarly, imagining a situation in which an individual loses an emotion-related job to a machine makes people feel more distressed compared to losing the ones that require advance cognitive abilities \cite{waytz2014botsourcing}. A study investigating the acceptance and use of eldercare robots has revealed that people are more likely to use robots if they already have positive attitudes towards robots and if they perceive robots’ minds as having less agency \cite{stafford2014does}. These findings suggest that people expect to see an artificial agent that have capabilities for both experience and agency; however, variance in these capacities influences people’s attitudes towards the agent. Research that explores the driving forces that can trigger mind attribution to artificial agents has shown that human-like appearance, social cues, and interaction setting can influence whether people attribute a mind to those agents. For example, a robot that has a face is perceived as having more agency and experience than a robot without a face \cite{broadbent2013robots}. Further, speed of movement can be used as a cue for perceiving mind in an entity \cite{morewedge2007timescale}. However, increasing human-like appearance does not necessarily increase mind perception; addition of human-like features increases the degree of mind perception only if the entities have already been classified as having minds \cite{martini2016seeing}. On the other hand, involvement in moral acts increases the likelihood of mind perception of an agent. Mind perception of a robot also increases when the robot is harmed \cite{ward2013harm}. Likewise, imagining a situation in which people treat a robot in a good way increases their likelihood of perceiving a mind in it \cite{tanibe2017we}.

\begin{figure*}
    \centering
    \includegraphics[width=0.99\textwidth]{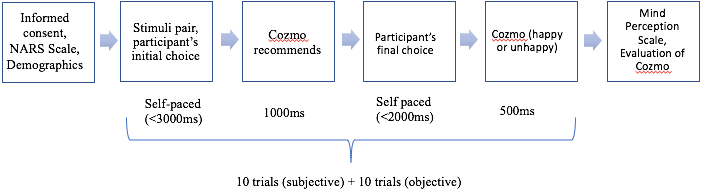}
    \caption{An example trial procedure. Tasks consisted of ten trials each.}
    \label{fig:proc}
\end{figure*}

Of particular importance to the present study, it has been found that attributing mind to an artificial agent may improve the process and consequences of the interaction \cite{wiese2012see,waytz2014botsourcing,tanibe2017we}.  Perceiving a mind in a robot influences the way in which people treat it.  Increased mind perception makes perceivers have more positive attitudes towards the robot \cite{tanibe2017we} and improves individual’s cognitive and joint-task performance with these agent \cite{wiese2012see}. Further, perceiving mental capacities in an autonomous vehicle increases its perceived trustworthiness, competence and responsibility \cite{waytz2014botsourcing}. Thus, it can be suggested that mind perception is an important factor in the use of robots in social settings, and studying it may help optimize human robot interactions, and improve consumers’ existing attitudes towards artificial agents. 

In line with the growing body of mind perception literature \cite{gray2007dimensions,gray2011more,gray2012feeling,waytz2010social}, this study aimed to explore the behavioral consequences of perceiving a mind in a robot. By manipulating the expressed agency and expressed experience of an autonomous robot that is positioned as a recommendation agent, we investigated the effects of experience and agency attribution on decisions.  We predict that a robot that displays agency and experience will have a greater influence on individuals’ preferences, compared to a robot that displays agency and experience to a lesser extent. Further, agency is expected to affect objective decisions more, where there is a correct answer, while experience is expected to have a bigger influence on subjective decisions. The key gap in the literature we intend to fill is whether attributing mind to a robot leads to conformity, which is operationalized as a decision change, upon interacting with it. Examining people’s tendency to conform after interacting with artificial agents that show variability in dimensions of mind, this study will shed light on the link between mind perception and conformity. Testing our predictions in both subjective and objective tasks will demonstrate whether informational and normative conformity apply to human-robot interaction. 

\section{Experiment}

\subsection{Participants}
A total of 49 (21 F, age range 18-25) students from a university campus community participated in this study in exchange for course credits. Approval was obtained from the institutional review board. 

\subsection{Material}
The experimental setting consisted of a screen for presenting the stimuli, an Anki Cozmo robot, two cubes of Cozmo that are used as buttons for making choices, and a laptop to control the experiment and record the results (\fig{fig:part_cozmo}).

\subsection{Procedure}
Before entering the experimental room, participants provided their informed consents and responded to the Negative Attitudes toward Robots Scale \cite{nomura2006experimental}. Then, participants received instructions, and provided demographic information that is to be fed in their recommendation agent Cozmo. Following this, they entered the experimental room and met Cozmo. After the experimenter and the participant performed the practice trial together, experimenter left the room and the experiment commenced. The participants either started with the subjective task and proceeded to the objective task, or the other way around. Task order was counterbalanced across participants. 

\begin{figure}
    \centering
    \begin{subfigure}[b]{0.99\columnwidth}
        \includegraphics[width=0.98\columnwidth]{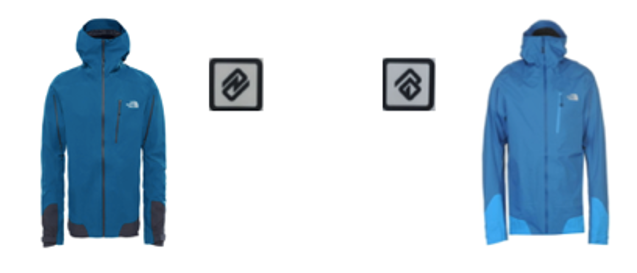}
        \caption{Subjective Task}
        \label{fig:objtask}
    \end{subfigure}
    \begin{subfigure}[b]{0.99\columnwidth}
        \includegraphics[width=0.98\columnwidth]{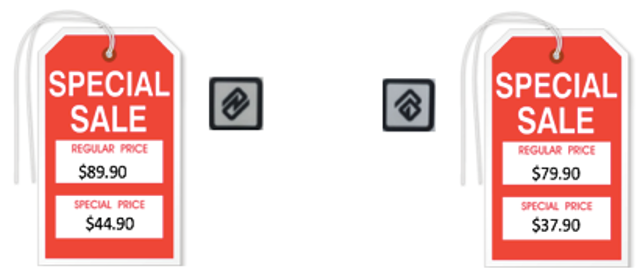}
        \caption{Objective Task}
        \label{fig:objtask}
    \end{subfigure}
    \caption{The subjective and the objective decision tasks. The participants are first asked to make a choice among the two items by tapping on a Cozmo cube, indicated by the symbol next to the item. Then after Cozmo acts, the participant chooses again.}
    \label{fig:tasks}
\end{figure}

For both tasks  questions appeared on a screen visible to Cozmo and the participant (\fig{fig:tasks}). Participants first responded to the question by tapping on one of the two cubes placed between them and Cozmo. On three, out of a total of ten trials, Cozmo agreed with the participant’s choice. On the remaining seven  critical trials Cozmo suggested the other option. On those trials, participants were asked to make a final choice, either maintaining their initial decisions, or changing them according to Cozmo’s suggestion. The trials occurred in a random order. 

In the subjective task, the questions asked the participants’ preference between colours or designs on the same clothing items. In the objective task, participants were asked to choose between two price tags the one that had a higher discount rate.  
 
\begin{figure}
    \centering
    \includegraphics[width=0.99\columnwidth]{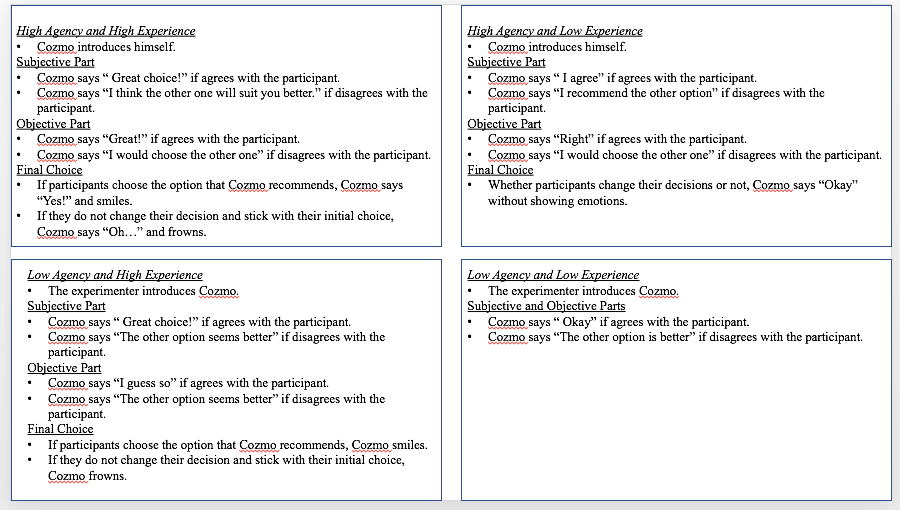}
    \caption{An example trial procedure. Tasks consisted of ten trials each.}
    \label{fig:cozmo_cond}
\end{figure}

The experiment involved a 2x2 factorial design, with between subject factors Agency (low, high) and Experience (low, high) . Participants were randomly assigned to one of four conditions. In each, Cozmo behaved in a different way with the aim of manipulating the degree to which participants’  attributed agency and experience to Cozmo (\fig{fig:cozmo_cond}). To control for a potential impact of the perceived gender of the robot, we did not use any words that can be related to gender.  

\subsection{Measures}
Two types of measures were used in this study: questionnaire data and experimental data which comprised the answers provided participants during their interaction with the robot. Our measure of interest was participants’ decision change after receiving robots’ suggestions in seven critical trials. These critical trials in both tasks demonstrated whether the participants’ tendency to conform changed depending on the nature of the decision.

Before interacting with the robot, participants in all conditions filled out the Negative Attitudes toward Robots Scale (NARS) in order to account for their existing attitudes towards interacting with robots \cite{nomura2006experimental}. This scale consists of three subscales measuring negative attitudes toward situations and interactions with robots, toward social interaction of robots, and emotions in interaction with robots. After completing the subjective and objective tasks, participants in all conditions filled out the Dimensions of Mind Scale that asked them to rate robot’s various mental capacities \cite{gray2007dimensions}. Consisting of two subscales (agency and experience), this scale measures participants perception of the robot’s capability of feeling pain and pleasure, of thinking and planning. It also serves as a manipulation check in our experimental design. 

At the end of the experiment, we asked participants to express whether they have perceived Cozmo’s gender as male or female. Plus, participants rated their overall experience and intention to use Cozmo in future. 

\begin{figure}
    \centering
    \includegraphics[width=0.99\columnwidth]{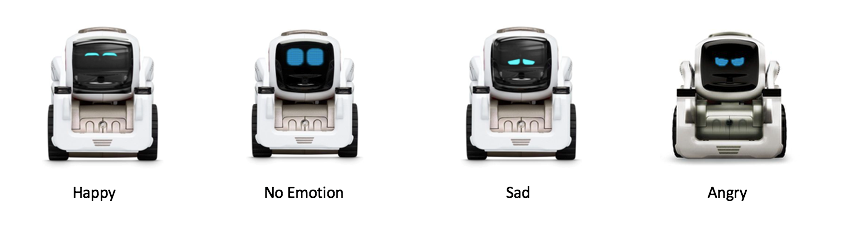}
    \caption{Faces Cozmo makes to convey emotions.}
    \label{fig:cozmo_face}
\end{figure}

\section{Results}
\begin{figure*}
    \centering
    \begin{subfigure}[b]{0.99\columnwidth}
        \includegraphics[width=0.98\columnwidth]{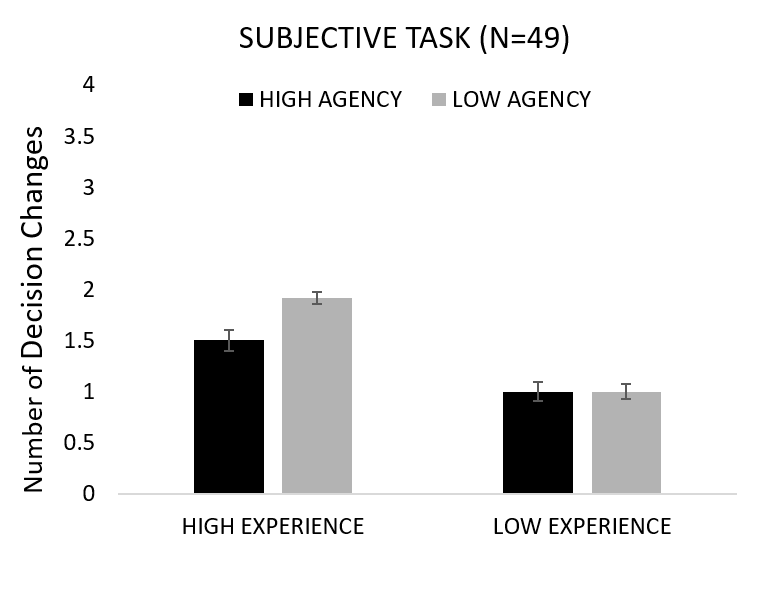}
        \caption{Subjective Task}
        \label{fig:st49}
    \end{subfigure}
    \begin{subfigure}[b]{0.99\columnwidth}
        \includegraphics[width=0.98\columnwidth]{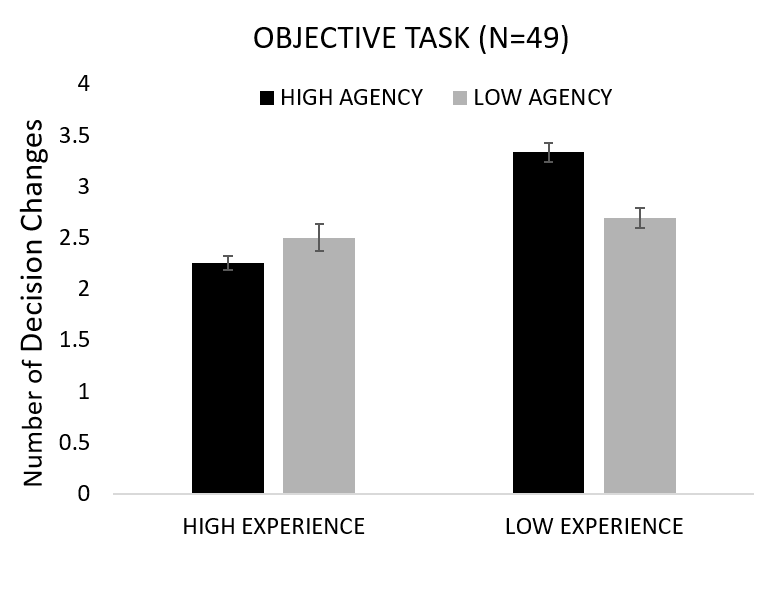}
        \caption{Objective Task}
        \label{fig:ot49}
    \end{subfigure}
    \caption{Average number of decision changes for each condition when $n=49$.}
    \label{fig:n49}
\end{figure*}

Our main goal was to explore whether participants conformed depending on the perceived agency and experience of the robot, and how the attributed levels of these dimensions influenced their tendency to conform in subjective and objective tasks separately. The collected data was analyzed using a 2 (Agency: high/low) x 2 (Experience: high/low) between subjects ANOVA with decision change as the dependent variable. We performed this analysis twice, once for each task (\fig{fig:st49} and \fig{fig:ot49}). 

The results revealed only a significant main effect of Experience on subjective decisions (F (1, 28) = 4.462), p < .05). The main effect of Agency, or the interaction were not found significant. Our agency manipulation was not a success, as the difference in the Dimensions of Mind scores between high and low agency conditions were not significant. This explains the lack of a main effect of Agency in the results. The significant main effect of Experience indicates that participants in the high experience condition changed their preference based subjective decisions more than those in the low experience condition. As hypothesized, a robot that displayed high levels of experience increased participants’ tendency to conform in the subjective task. The 2x2 ANOVA with the objective task revealed a similar pattern of results, however neither the main effects or the interaction reached significant. We observed a trend between the low and high experience conditions, such that participants in the low experience condition were more likely to change their objective decisions compared to the high experience condition. 
 
\begin{figure*}
    \centering
    \begin{subfigure}[b]{0.99\columnwidth}
        \includegraphics[width=0.98\columnwidth]{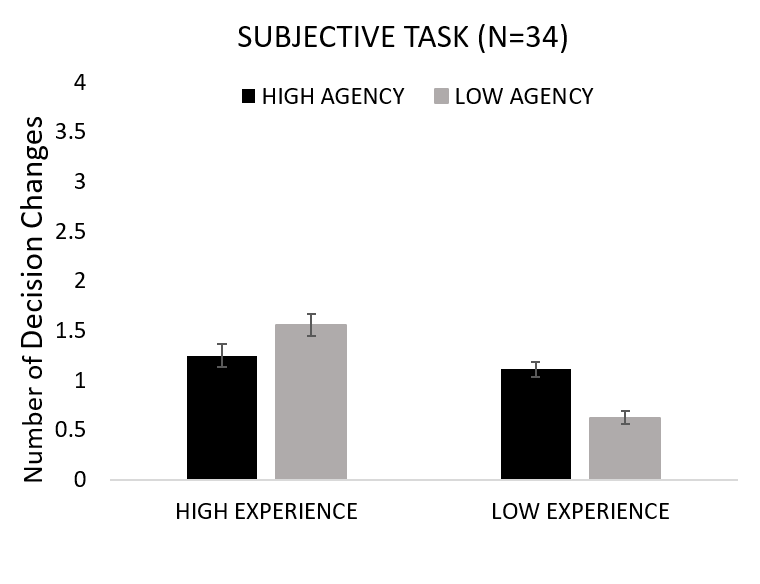}
        \caption{Subjective Task}
        \label{fig:st34}
    \end{subfigure}
    \begin{subfigure}[b]{0.99\columnwidth}
        \includegraphics[width=0.98\columnwidth]{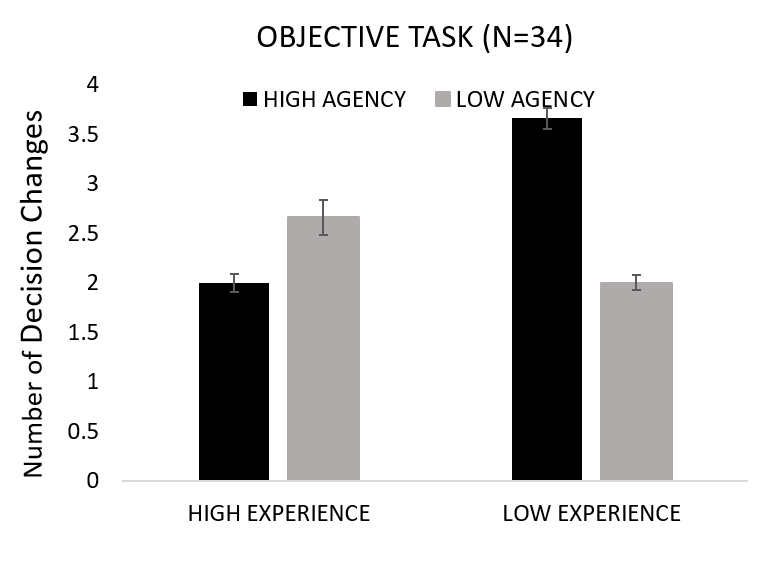}
        \caption{Objective Task}
        \label{fig:ot34}
    \end{subfigure}
    \caption{Average number of decision changes for each condition when $n=34$, corresponding to the ANOVA results with the expected agency and experience used as covariate.}
    \label{fig:n34}
\end{figure*}

During the data collection process, we realized that participants already had some preconceptions about Cozmo in terms of its agency. For instance, when they first saw Cozmo, they made eye contact and talked to Cozmo. They seemed as if they were trying to find cues to validate their expectations. After observing such a tendency, we decided to measure participants initial expectations in an effort to control for their possible impact on levels of conformity. We asked the remaining 34 participants to rate what they expected from a robot in terms of capacity to plan and act, and sense and feel on a 10- point scale. Participants expected robots to have high levels of Agency (mean= 6.65, SD=2.24), and low levels of Experience (mean=3.29, SD=2.1). We used these two variables as covariates and repeated the ANOVA analyses (\fig{fig:st34} and \fig{fig:ot34}). In the subjective task, neither the main effects, nor the interaction was significant. However, in the objective task the analysis yielded a significant interaction of Experience and Agency ($F(1,28)=12.363, p<.005$), although the main effects did not reach significance. 

To explore the nature of the interaction effect, we performed pairwise comparisons. There was a significant difference between low agency- low experience and high agency-low experience conditions (t(15)=-3.131,  $p<.01$). This suggests that perceived agency had an influence on participants’ willingness to change their objective decisions and conform to Cozmo in the low experience condition. The difference in decision change between high agency-high experience and high agency-low experience conditions was also found significant (t(15)=-2.97, $p<.01$), indicating that level of experience negatively influenced individuals’ tendency to conform in an objective task when they attributed high agency to Cozmo. 

To summarize, in the subjective task, by displaying high levels of experience, Cozmo induced decision change in the participants. As for the effect of perceived agency we did not observe a significant impact on decision change. This is likely because of our participants high agency expectations at the outset of the experiment, which possibly blunted our manipulation. However, when the initial expectations were controlled, we observed an interaction between high agency and low experience in the objective task. Here, by displaying high levels of agency and low levels of experience, Cozmo induced decision change in the participants.

\section{DISCUSSION AND LIMITATIONS}
Exploring the behavioral consequences of mind perception in human-robot interaction, the current study provides useful insights for understanding social robotics. Firstly, it analyzes the nature and outcomes of this interaction in a real-life setting. Secondly, previous studies studied conformity in terms of group dynamics, either focusing on subjective or objective decisions. Investigating individuals' tendency to conform to a single robot while making objective and subjective decisions under the same conditions, this study forms a basis for understanding the underlying dynamics of decision making with robots, and provides a new aspect that should be investigated in future. 

Our findings also show that humans’ existing expectations from robots should be taken into account while developing robots that can address to a wide range of people and designing interaction settings. As their expectation influences the consequences of the interaction, researchers should account for this factor which is mostly neglected in the field. Further, people may not always be honest with their thoughts and feelings. In this study, by challenging, protesting, and arguing with Cozmo, people showed that they indeed perceived Cozmo as a mindful agent. However, they were reluctant to acknowledge Cozmo’s mind, as indicated by the ratings in the Dimensions of Mind Scale. The perceived level of agency and experience and attribution of mind that happened during the interaction could not be observed in the evaluation that is made afterwards. This finding indicates that self-report measures may not always reflect the reality, as individuals may have a tendency to disguise their ideas in social studies. 

There are a few limitations of this study. First of all, we were unable to successfully manipulate perceived levels of agency. During the data collection process, we realized that people had already attributed agency to robots, and did not decrease the level of agency that they ascribed to Cozmo regardless of the condition. Thus, manipulating a factor that is already perceived as high can be a drawback in studying mind perception in robots. The problems that are encountered in this study should be taken into account in future studies that dwell upon social side of robotics. 

\section{CONCLUSION AND FUTURE RESEARCH}
Autonomous robots have been introduced to us in the last decade and they are expected to spread across different realms of our daily lives by 2030 \cite{stone2016artificial}. By embedding the theoretical framework of mind perception into the human robot interaction in a shopping context, the current study provides insights regarding autonomous robots’ influence on individuals’ decisions. By exploring the link between mind perception and decision change, it has also extended our knowledge about the behavioral consequences of mind perception in an interactive setting. 

Continuous interaction with mindful entities that have high ratings of agency and competence may decrease people’s confidence in their decisions. It may lead to underconfidence in the long run and increase humans’ dependence on robots, which will impair their ability to distinguish between right and wrong decisions of robots and detect robots’ errors. As they prioritize robots’ suggestions, their feedback processing can be biased, too. Thus, people’s confidence judgments, error monitoring and feedback processing after interacting with an artificial agent should be studied further.

As the number of interactions with autonomous agents increases day by day, individual and situational factors that can encourage or discourage people to interact with robots should be studied further. Research has shown that interacting with an agent that is perceived as having a mind improves the quality of the interaction \cite{wiese2012see,waytz2014botsourcing,tanibe2017we}.

However, people may also desire to interact with mindless agents in various contexts, such as purchasing sin products and consulting about private issues. Moreover, they may expect agents to demonstrate high levels of agency and low levels of experience in certain contexts such as financial interactions, medical and statutory advising, while they may expect agents to signal more experience than agency in teaching, psychiatry, and various fields of arts. Exploring goal-oriented interactions with robots and personality variables can improve the design and use of these artificial agents, increase the number of people who are willing to interact with them, and improve the nature of these interactions.

\addtolength{\textheight}{-5cm}  
\bibliographystyle{IEEEtran}
\bibliography{references}

\end{document}